%% file: arxiv.tex
\documentclass{article}

\usepackage{iclr2027_conference,times}
\input{math_commands.tex}

\usepackage{hyperref}
\usepackage{url}
\usepackage{booktabs}
\usepackage{tabularx}
\usepackage{array}
\usepackage{graphicx}
\usepackage{xcolor}
\usepackage{colortbl}
\definecolor{autocompactrow}{HTML}{F2F5FA}
\usepackage{tikz}
\usepackage{pgfplots}
\usetikzlibrary{arrows.meta,positioning,calc,fit,backgrounds}

\definecolor{ackeep}{HTML}{5873C7}
\definecolor{acsummary}{HTML}{408A82}
\definecolor{acdrop}{HTML}{D56A4A}
\definecolor{acneutral}{HTML}{4B5563}
\pgfplotsset{compat=1.18}
\newcolumntype{Y}{>{\raggedright\arraybackslash}X}

\title{AutoCompact: Learning When to Compact\\Context in Long-Horizon Coding Agents}

\author{Xuan Zhang\thanks{Equal Contribution}~$^{~1}$, Longtao Zheng$^{*2}$, Cunxiao Du$^1$~, Bo An$^2$, Xin Dong$^3$\\
$^1$Singapore Management University
$^2$Nanyang Technological University
$^3$Harvard University \\
\texttt{\{xuanzhang826, ltzheng01\}@gmail.com} \\
}

\iclrfinalcopy

\begin{document}

\maketitle

\begin{abstract}
Coding agents solve repository-level software engineering tasks through long trajectories of code inspection, search, editing, and testing. As a task progresses, earlier exploration becomes stale, so managing context is more than avoiding overflow: an agent must decide when to compact, what working state to preserve, and how to continue from it. We introduce AutoCompact, which trains a coding agent to make these decisions as part of its policy. To collect training data, we run the base agent on coding tasks and use a judge to review its compaction decisions, summaries, and actions after compaction. Flawed outputs are replaced with corrected ones before being executed in the environment, so each trajectory continues from the corrected decisions. We use these trajectories for supervised fine-tuning, then jointly optimize coding and compaction through reinforcement learning with task-success rewards. Experiments on SWE-bench Verified and SWE-PolyBench Verified show that AutoCompact improves pass rates over the base model by an absolute 9.2\% and 5.0\%, respectively. The improvements hold across all evaluated inference budgets, with a 256K context window that never overflows and with a 16K window whose overflow triggers fallback compaction.
\end{abstract}

\section{Introduction}
\label{sec:intro}
Large language model (LLM) agents have demonstrated strong capabilities on repository-level software engineering tasks by inspecting code, searching the repository, editing files, and running tests~\citep{swebench,sweagent,openhands}. Resolving complex issues often requires investigating multiple candidate causes and iterating between implementation and validation, so trajectories accumulate code snippets, tool outputs, and intermediate findings. As the agent moves from one stage of a task to the next, much of this information, such as exploratory hypotheses, failed attempts, and detailed tool outputs, becomes stale. Keeping the full trajectory thus fills the context with outdated details, whereas the next stage needs only a compact \emph{working state}: the conclusions reached so far, the relevant code and workspace status, and the remaining actions. Effective context management therefore requires an agent to decide \emph{when} to compact, \emph{what} information to preserve in the working state, and \emph{how} to continue execution from it.

Existing approaches address only part of this problem. Length-triggered methods such as CompactionRL~\citep{compactionrl} compact only when the remaining context budget falls below a fixed threshold, which ties compaction to context length rather than task progress: stale exploration accumulates until the threshold is reached, and compaction may then occur in the middle of an unresolved stage whose evidence is still needed. Proactive methods instead let the agent decide when to compact, either through an inference-time rubric~\citep{selfcompact} or through fine-tuning on trajectories into which compression calls are inserted offline~\citep{contextasatool}. However, deciding when to compact is not sufficient: the working state may omit or misrepresent key information, and even when it is correct, the agent may fail to follow it, revisiting completed exploration or ignoring the intended next action. Neither kind of proactive method addresses these failures: a rubric provides no training signal, and offline insertion keeps the original actions after each inserted call.

We introduce AutoCompact, which trains a coding agent to decide when to compact, what to preserve, and how to continue as part of its policy. To collect training data, we run the base agent on training tasks and use a judge to review its compaction decisions, the summaries it writes, and its actions after compaction. The judge replaces flawed outputs with corrected ones before they are executed in the environment, so each trajectory continues from the corrected decisions and shows how to act after compaction. We use these trajectories for supervised fine-tuning (SFT), and then apply outcome-based reinforcement learning (RL), which jointly optimizes coding and compaction with binary task success as the only reward signal.

We evaluate AutoCompact on SWE-bench Verified~\citep{swebench,swebenchverified} and SWE-PolyBench Verified~\citep{swepolybench}, achieving pass rates of 39.6\% and 24.5\%, respectively, with improvements of 9.2\% and 5.0\% over the base model. We further examine cost efficiency on SWE-bench Verified under two complementary context regimes: a 256K setting in which no trajectory reaches the forced-compaction threshold, and a constrained 16K setting in which all methods share the same length-triggered fallback. The trained policies consistently improve performance across all evaluated budgets from \$0.10 to \$4.00 per task, with outcome-based RL providing further gains over SFT, especially at low budgets. Comparing the same trained checkpoint with and without executing compaction further shows that compaction is particularly beneficial under tight inference budgets, supporting its contribution to efficient task completion.

\begin{figure}[t]
\centering
\includegraphics[width=\linewidth]{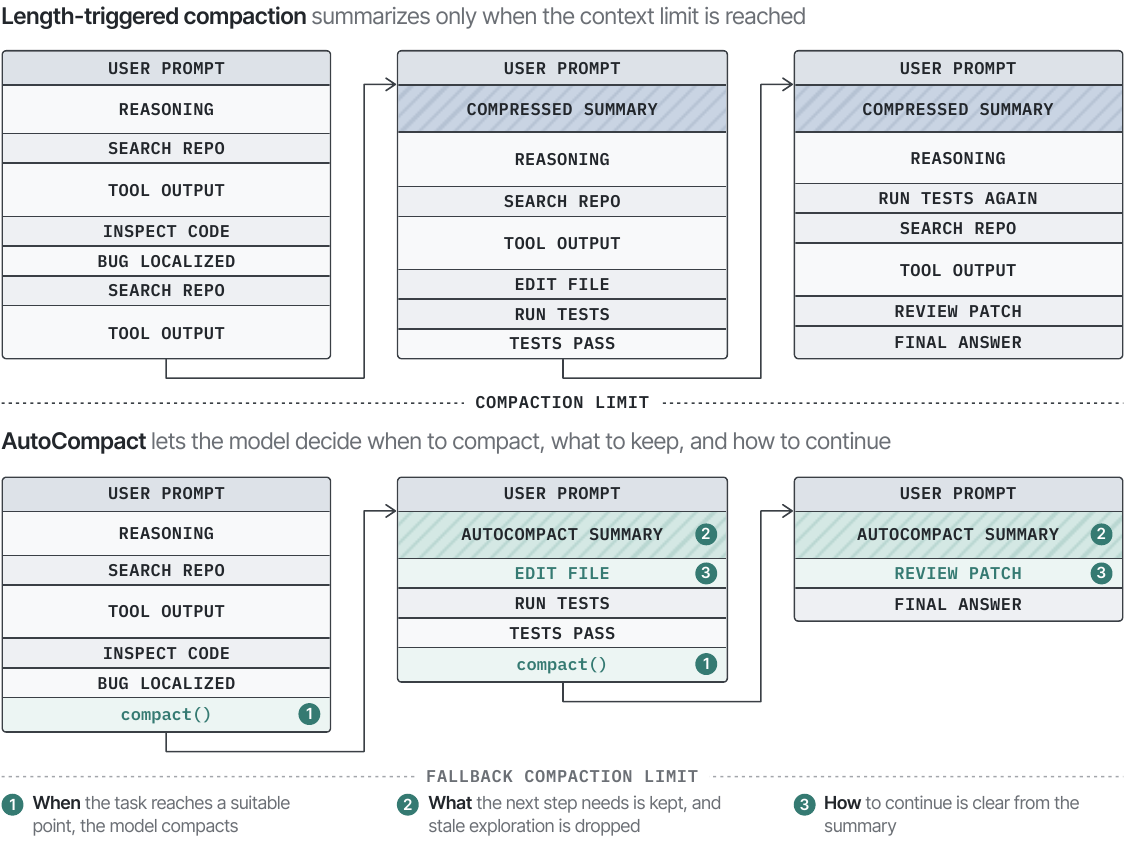}
\caption{Length-triggered compaction versus AutoCompact. Under length-triggered compaction, the agent keeps accumulating context after the bug is localized, such as repeated searches with long outputs, and summarizes only when the context limit is reached. AutoCompact lets the model decide (1) when to compact, (2) what to keep, rewriting useful findings into a working-state summary while dropping stale exploration, and (3) how to continue from it.}
\label{fig:mechanism}
\end{figure}

Our main contributions are as follows. (1) We introduce AutoCompact, which corrects compaction decisions, summaries, and subsequent actions during data collection and executes the corrected outputs in place of the original ones. (2) We collect 1,052 judge-corrected trajectories on SWE-rebench~\citep{swerebench} to initialize the policy via SFT, followed by outcome-based RL that jointly optimizes coding and compaction without compaction-specific reward shaping. (3) We demonstrate improved task success on SWE-bench Verified and SWE-PolyBench Verified, with gains across inference budgets and context settings. To rule out that these gains stem from training alone, we ignore the compaction calls of the same trained model; the resulting drop in pass rates confirms that executing compaction itself contributes.

\section{Method}
\label{sec:method}

AutoCompact equips a coding agent with a \texttt{compact()} action (Section~\ref{sec:inference_compaction}) and trains the policy in two stages. We first collect on-policy trajectories with judge-guided corrections to compaction timing, the generated working state, and the actions immediately following compaction (Section~\ref{sec:judge_sft}). These corrected trajectories provide the training data for SFT. We then apply outcome-based RL to jointly optimize coding and compaction using final task success as the reward (Section~\ref{sec:rl}).

\subsection{Proactive Context Compaction}
\label{sec:inference_compaction}

\begin{figure}[t]
\centering
\includegraphics[width=\linewidth]{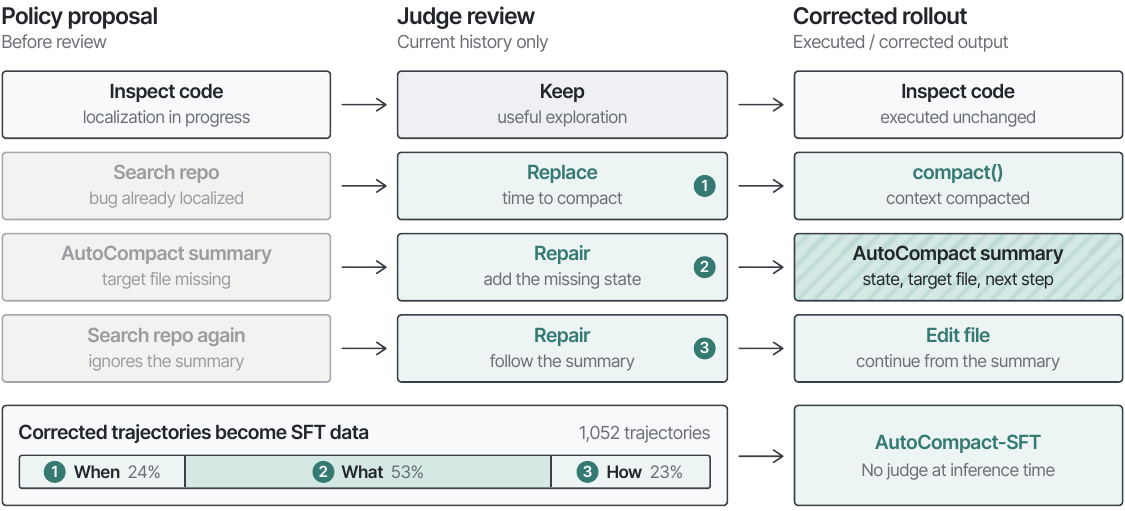}
\caption{Judge-guided on-policy data collection. At each step, the judge reviews the policy's proposal using only the current execution history. Useful exploration is kept, while corrections target (1) when to compact, e.g., replacing a search after the bug is localized with \texttt{compact()}; (2) what to keep, e.g., repairing a summary that omits the target file; and (3) how to continue, e.g., redirecting a repeated search toward the edit planned in the summary. The environment executes the corrected outputs, and the resulting trajectories supervise SFT; the trained agent needs no judge.}
\label{fig:annotation_pipeline}
\end{figure}

We augment the agent with a \texttt{compact()} action that can be invoked proactively during execution. The agent otherwise operates normally by inspecting files, editing code, and running tests. When it determines that the current stage has been sufficiently resolved and the accumulated history can be summarized, it may invoke \texttt{compact()} before reaching the context limit. For example, the agent may compact after localizing the cause of an issue and before proceeding to implementation.

Compaction replaces earlier interaction history with a model-generated working-state summary headed by \texttt{\# Auto Context Summary}, while retaining the original task unchanged. The summary is intended to preserve established findings, relevant code and workspace state, and remaining actions, while omitting unnecessary details of prior exploration. The agent then resumes execution from the resulting context, as illustrated in Figure~\ref{fig:mechanism}.

This mechanism makes compaction a model decision rather than solely a response to context-window pressure. Its effectiveness therefore depends on three behaviors: choosing an appropriate compaction point, constructing an accurate working state, and continuing reliably from that state. We next describe how we supervise these behaviors through judge-guided on-policy data collection.

\subsection{Judge-Guided Data Collection and SFT}
\label{sec:judge_sft}

Prompting alone does not induce proactive compaction. In our preliminary experiments, the base model rarely invoked \texttt{compact()} before the context limit, even when compaction rules were specified in the agent prompt. Proactive compaction introduces a behavior that is uncommon in the base policy: the model must interrupt an otherwise valid execution trajectory and replace its accumulated context before any context-limit constraint forces it to do so. We therefore collect training trajectories with online judge corrections and use them for SFT. These corrections provide demonstrations of compaction and subsequent execution that the base policy rarely produces on its own. The judge is used only during data collection; the trained agent runs without the judge at inference time.

To collect such demonstrations, we roll out the base policy on software engineering tasks and use a judge to review each proposed action as execution proceeds. The judge receives only the trajectory history available at the current step and determines whether the proposed action is satisfactory, following a predefined annotation protocol that specifies appropriate compaction points, the information that should be preserved in the compacted working state, and how post-compaction behavior should continue. Figure~\ref{fig:annotation_pipeline} provides an overview of this online corrective data-collection process, with one concrete example of each of the three types of correction described in the next paragraph.

We supervise three types of decisions corresponding to the main failure modes of compaction. \textit{Trigger correction} checks whether the current point is appropriate for compaction: the agent should compact when the current stage has been sufficiently resolved, but continue exploring when intermediate evidence is still needed. \textit{Working-state correction} checks whether the generated \texttt{\# Auto Context Summary} accurately preserves the information needed for subsequent execution, including relevant conclusions, code and workspace state, and remaining actions. \textit{Continuation correction} checks whether the first actions after compaction follow the resulting working state, rather than revisiting completed exploration or ignoring the intended next step, such as re-running a search whose result is already recorded in the working-state summary.

Corrections are applied before execution, so each one shapes the rest of the trajectory. When the judge identifies an output unsatisfactory, it writes a corrected output, which the environment executes in place of the original; the policy then generates the following actions from this corrected history. This differs from post-hoc annotation, where corrections do not influence later states.

Finally, we fine-tune the base policy on the corrected trajectories with the standard next-token prediction objective. The resulting policy, AutoCompact-SFT, learns both ordinary coding behavior and the three compaction behaviors introduced above: selecting appropriate compaction points, constructing actionable working states, and reliably continuing execution afterward. SFT provides a behavioral initialization for proactive compaction, but it optimizes the locally corrected decisions rather than their eventual effect on task completion. We then further optimize the policy using outcome-based RL, which directly targets task success.

\subsection{Joint RL for Task Solving and Compaction}
\label{sec:rl}

Starting from the SFT checkpoint, we further train the agent on SWE-Gym~\citep{swegym} with end-to-end multi-turn RL~\citep{simpletir}, using group relative policy optimization (GRPO)~\citep{deepseekmath} and a fully asynchronous RL system. Each rollout receives a binary outcome reward based on whether the final patch passes the task tests. Within each rollout group, GRPO converts these outcomes into relative advantages, which provide a shared learning signal for all model-generated tokens in a trajectory, including regular coding actions, the decision to invoke \texttt{compact()}, the generated context summary, and the continuation after compaction.

Compaction therefore requires neither a separate auxiliary objective nor compaction-specific reward shaping. Instead, task solving and context management are trained jointly under the final task reward, allowing the policy to learn when compaction is beneficial, what information should be preserved, and how execution should continue afterward.

Compaction also changes how trajectories are converted into training sequences, because each call to \texttt{compact()} rewrites the context. As a result, a trajectory with compaction no longer forms a single sequence in which each model turn extends the previous context. We therefore split each trajectory into segments wherever the context prefix is rewritten, such that within each segment every model turn is conditioned on a prefix that only grows. All segments of a trajectory share its advantage, so the decision to compact, the summary, and the actions after compaction receive the same outcome-based signal even though they fall into different segments. The policy loss is averaged at the token level over all segments in a batch, without a KL penalty or an entropy bonus, so the binary outcome reward is the only training signal for both coding and compaction.

\section{Experiments}
\label{sec:experiments}

Our experiments address three questions. First, does AutoCompact improve task success over full-history execution and over existing length-triggered and proactive compaction methods (Section~\ref{sec:results})? Second, how do proactive compaction, outcome-based RL, and the execution of compaction itself contribute across inference budgets and context settings (Section~\ref{sec:analysis})? Third, how do the summaries produced after SFT and after RL differ (Section~\ref{sec:case-study})?

\subsection{Experimental Setup}
\label{sec:setup}

\textbf{Models and scaffold.} We use Qwen3-Coder-30B-A3B-Instruct~\citep{qwen} as the base model throughout our experiments and evaluate all model variants on two repository-level software-engineering benchmarks: SWE-bench Verified~\citep{swebench,swebenchverified} and SWE-PolyBench Verified~\citep{swepolybench}. All variants use the same terminal-based read-eval-print loop (REPL) scaffold~\citep{ttyuse, annotation}, augmented with the model-invocable \texttt{compact()} action described in Section~\ref{sec:inference_compaction}. During supervised data collection, we additionally use GPT-5.5-Codex as the judge for online correction; the judge is removed during evaluation.

\textbf{Training.} For SFT, we roll out the base model on 379 SWE-rebench tasks~\citep{swerebench} and collect 1,052 judge-corrected trajectories using the procedure in Section~\ref{sec:judge_sft}, after filtering out malformed requests, summary loops, and off-track continuations. Of these trajectories, 24\% supervise compaction triggers, 53\% supervise working-state construction, and 23\% supervise post-compaction continuation. We train for two epochs with a learning rate of $5\times10^{-7}$ and a batch size of 8. Starting from AutoCompact-SFT, we further optimize the policy with GRPO on SWE-Gym~\citep{swegym}. For each task, we sample 8 trajectories and assign a binary outcome reward according to whether the final patch passes the task-specific test suite. RL uses a learning rate of $1\times10^{-6}$ and a batch size of 64, with rollouts capped at 50 environment steps and 32K tokens.

\begin{figure}[!t]
\centering
\includegraphics[width=\linewidth]{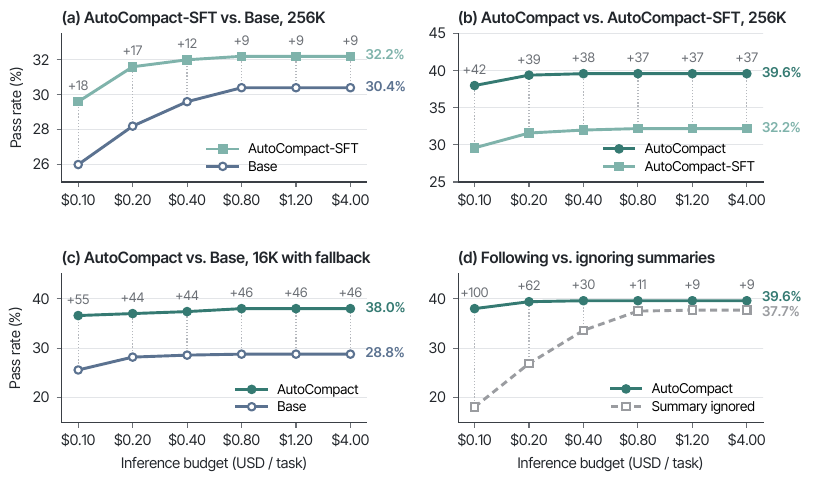}
\caption{Pass rate versus inference budget on SWE-bench Verified. (a) AutoCompact-SFT versus Base at 256K, where no trajectory reaches the forced-compaction threshold. (b) AutoCompact (after RL) versus AutoCompact-SFT at 256K. (c) AutoCompact versus Base at 16K, with a shared forced-compaction fallback. (d) The same checkpoint following versus ignoring summaries. Budget categories are equally spaced; gains denote additional solved tasks per 500 tasks.}
\label{fig:budget}
\end{figure}

\textbf{Evaluation.} We report pass rates on SWE-bench Verified and SWE-PolyBench Verified using full agent runs without an inference-cost cap. All reported quantitative evaluation results are averaged over three runs. To assess cost efficiency, we additionally evaluate SWE-bench Verified under six per-task inference budgets from \$0.10 to \$4.00 (Figure~\ref{fig:budget}). Each rollout is terminated when its cumulative inference cost reaches the specified budget. We estimate inference costs from token usage using the official Alibaba Cloud Model Studio pricing for Qwen3-Coder-30B-A3B-Instruct~\citep{alibabapricing}. To account for prefix caching, we price cached tokens at 20\% of the standard input-token rate, following Alibaba Cloud Model Studio's implicit-caching pricing, and price uncached input and output tokens at their respective standard rates.

\begin{table*}[t]
\centering
\caption{Full-run pass rates without an inference-cost cap, grouped by compaction trigger. All methods share the same base model and scaffold. In the Context column, 256K denotes the model context window, while $\dagger$ marks a 16K forced-compaction threshold.}
\label{tab:main-results}
\begingroup
\small
\setlength{\tabcolsep}{3.5pt}
\renewcommand{\arraystretch}{1.08}
\begin{tabular}{l c c c c c}
\toprule
\textbf{Method}
& \textbf{Trigger}
& \textbf{Optimization}
& \textbf{Context}
& \begin{tabular}[c]{@{}c@{}}\textbf{SWE-bench}\\\textbf{Verified (\%)}\end{tabular}
& \begin{tabular}[c]{@{}c@{}}\textbf{SWE-PolyBench}\\\textbf{Verified (\%)}\end{tabular} \\
\midrule
\rowcolor{black!6}
\multicolumn{6}{l}{\rule{0pt}{10pt}\textbf{Full-history baseline}} \\
Base & \textemdash & \textemdash & 256K & 30.4 & 19.5 \\
\addlinespace[1.5pt]
\rowcolor{black!6}
\multicolumn{6}{l}{\rule{0pt}{10pt}\textbf{Length-triggered compaction}} \\
Fixed Compaction & Length & \textemdash & 16K$^{\dagger}$ & 28.8 & 18.6 \\
CompactionRL~\citep{compactionrl} & Length & RL & 16K$^{\dagger}$ & 32.7 & 19.8 \\
\addlinespace[1.5pt]
\rowcolor{black!6}
\multicolumn{6}{l}{\rule{0pt}{10pt}\textbf{Proactive compaction}} \\
SelfCompact~\citep{selfcompact} & Rubric & \textemdash & 256K & 31.7 & 20.6 \\
SWE-Compressor~\citep{contextasatool}& Learned & SFT & 256K & 31.0 & 20.1 \\
\rowcolor{autocompactrow}
AutoCompact-SFT & Learned & SFT & 256K & 32.2 & 21.7 \\
\rowcolor{autocompactrow}
AutoCompact & Learned & SFT $\rightarrow$ RL & 256K & \textbf{39.6} & \textbf{24.5} \\
\bottomrule
\end{tabular}
\endgroup
\end{table*}

\textbf{Baselines.} We compare AutoCompact against several context-management strategies. Base uses the same underlying model and scaffold without learned proactive compaction. Fixed Compaction triggers compaction when the remaining context budget falls below a predefined threshold. SelfCompact~\citep{selfcompact} uses an inference-time rubric to guide compaction decisions, while CompactionRL~\citep{compactionrl} jointly trains task execution and summary generation under a length-triggered compaction mechanism. SWE-Compressor~\citep{contextasatool} learns to invoke a context-management tool through supervised fine-tuning on trajectories reconstructed offline by inserting compression calls. For this SFT comparison, we collect trajectories from comparable SWE-rebench task sets and use similar numbers of training trajectories for SWE-Compressor and AutoCompact-SFT. Unlike these baselines, AutoCompact corrects the agent's compaction decisions and subsequent actions during data collection, directly supervising how the agent uses the compacted state.

\subsection{Main Results}
\label{sec:results}

As shown in Table~\ref{tab:main-results}, AutoCompact achieves the highest full-run pass rates on both benchmarks, improving over Base by 9.2\% on SWE-bench Verified and 5.0\% on SWE-PolyBench Verified.

\textbf{Length-triggered compaction yields limited gains.} Fixed Compaction underperforms Base by 1.6\% and 0.9\% on the two benchmarks, indicating a performance cost in this constrained-context setting. CompactionRL improves upon Fixed Compaction by 3.9\% and 1.2\% through RL. However, its gains over Base remain limited, leaving a substantial gap to AutoCompact.

\textbf{Proactive compaction outperforms full-history execution.} Under the 256K setting, all evaluated proactive methods, including SelfCompact, SWE-Compressor, and the AutoCompact variants, outperform Base on both benchmarks. SelfCompact achieves improvements through inference-time guidance alone, while SWE-Compressor learns compaction-tool use through SFT. These results suggest that proactive context management helps even when the full history fits in context.

\textbf{Online judge-guided SFT and outcome-based RL both add gains.} AutoCompact-SFT exceeds SWE-Compressor by 1.2\% and 1.6\% on the two benchmarks. Since both use the same base model, scaffold, and comparable training data, this comparison suggests that online judge-guided correction is more effective than offline insertion into completed trajectories. AutoCompact-SFT learns from trajectories in which the corrected outputs, including corrected actions after compaction, were actually executed, whereas SWE-Compressor inserts compaction calls into completed trajectories and keeps their original subsequent actions. Outcome-based RL then improves pass rates by a further 7.4\% and 2.8\% over AutoCompact-SFT on the two benchmarks.

\subsection{Further Analysis}
\label{sec:analysis}

\textbf{Proactive compaction helps before context overflow.} We first compare Base with AutoCompact-SFT under the 256K long-context setting. Because no trajectory reaches the forced-compaction threshold, this comparison tests whether supervised training for proactive compaction improves agent performance even when the full interaction history fits within the context window. As shown in Figure~\ref{fig:budget}~(a), AutoCompact-SFT achieves a higher pass rate than Base across all evaluated inference budgets. This result shows that retaining the complete interaction history is not necessarily optimal, even when the context window is sufficiently large. Over long trajectories, obsolete hypotheses, failed attempts, and verbose tool outputs can accumulate and distract the model from the current working state. By replacing stale history with a concise working state, AutoCompact-SFT improves performance before context becomes a bottleneck.

\textbf{Outcome-based RL improves beyond supervised imitation.} Under the same 256K protocol, we compare AutoCompact-SFT with AutoCompact. As shown in Figure~\ref{fig:budget}~(b), AutoCompact consistently outperforms AutoCompact-SFT across all inference budgets, with the largest gains in the low-budget regime. These results suggest that SFT establishes the basic compaction behavior, while outcome-based RL further improves how this behavior is used to maximize task success. Since low budgets cut rollouts short, the larger gains at these budgets indicate that RL makes the agent more cost-efficient rather than simply relying on longer trajectories.

\begin{figure}[t]
  \centering
  \begin{minipage}{1\linewidth}
    \centering\small
    \includegraphics[width=\linewidth]{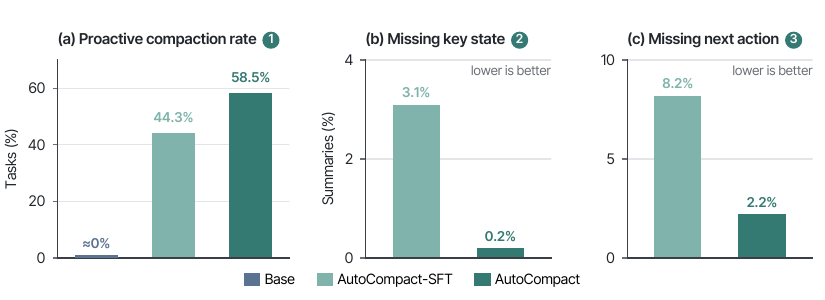}
    \caption{Compaction behavior, with (1), (2), and (3) marking when to compact, what to keep, and how to continue (Figure~\ref{fig:mechanism}). (a) Fraction of tasks in which the model proactively invokes \texttt{compact()}. (b) Fraction of summaries that omit relevant task or workspace state. (c) Fraction of summaries that omit a concrete next action. Lower is better in (b) and (c).}
    \label{fig:compaction-behavior}
  \end{minipage}
\end{figure}

\textbf{RL improves compaction behavior without compaction-specific rewards.} Figure~\ref{fig:compaction-behavior} compares compaction behavior before and after RL. While Base rarely invokes \texttt{compact()}, AutoCompact-SFT already uses it on 44.3\% of tasks, indicating that supervised training establishes the basic behavior. RL extends its use to 58.5\% of tasks while reducing both types of summary omission: key-state omissions fall from 3.1\% to 0.2\%, and next-action omissions fall from 8.2\% to 2.2\%. These statistics are computed on trajectories for all 500 SWE-bench Verified tasks using keyword-based screening, supplemented by random manual spot checks. The two trends are complementary: compaction is used across more tasks, yet fewer summaries are flagged for missing state or next-action information. This suggests that task-success rewards encourage the agent to preserve information needed to continue the task from the compacted state.

\textbf{Proactive compaction complements a length-triggered fallback.} We evaluate Base and AutoCompact under a shared 16K forced-compaction threshold. This setting represents a context-constrained runtime in which both systems use forced compaction as a fallback, while AutoCompact can also invoke \texttt{compact()} proactively before reaching the threshold. As shown in Figure~\ref{fig:budget}~(c), AutoCompact achieves a higher pass rate than Base across all evaluated inference budgets. This result shows that learned proactive context management remains effective when deployed together with a length-triggered fallback, allowing the agent to reorganize its context before compaction becomes mandatory. It also indicates that compaction based on task progress is more effective than compaction based solely on the length of the context.

\textbf{Compaction contributes to the gains of the trained agent.}\label{sec:same-checkpoint} The gains of AutoCompact may reflect both better agent behavior learned during training and the use of compaction during inference. To assess the latter contribution, we compare normal execution with a Summary-ignored variant of the same trained checkpoint. In this variant, \texttt{compact()} calls are skipped: no summary is generated, and execution continues with the existing history. Both conditions retain all learned parameters, allowing us to test whether executing compaction provides a benefit beyond the improvements encoded in those parameters. Figure~\ref{fig:budget}~(d) shows that normal execution achieves higher pass rates, with the largest advantage under tight inference budgets. This indicates that the observed gains cannot be attributed solely to better agent behavior learned during training: part of the benefit depends on actually executing compaction. The advantage is 19.9\% at \$0.10 and remains 1.9\% at \$4.00, so executing compaction improves cost efficiency without reducing final performance.

\begin{figure}[t]
  \centering
  \begin{minipage}{\linewidth}
    \centering\small
    \includegraphics[width=\linewidth]{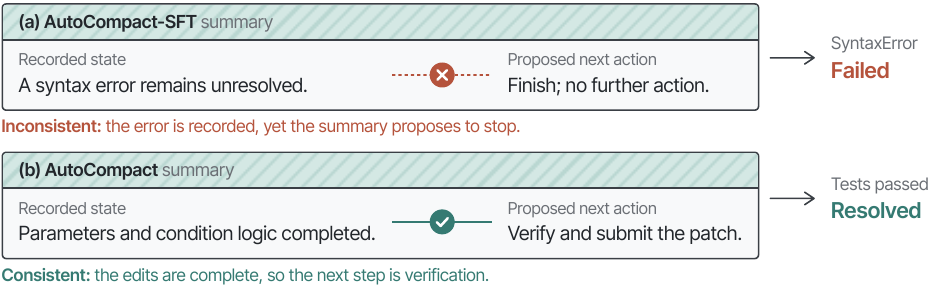}
    \caption{Summary self-consistency on SWE-bench Verified task \texttt{django-13809}. This qualitative example contrasts paraphrased summaries from AutoCompact-SFT and AutoCompact with their final evaluation outcomes. Each summary records a state and proposes a next action; the marker between them indicates whether the action is consistent with the recorded state.}
    \label{fig:self-consistency-case}
  \end{minipage}
\end{figure}

\subsection{Case Study: Summary Self-Consistency under SFT and RL}
\label{sec:case-study}

Preserving the relevant state and specifying a next action do not by themselves make a summary usable. The summary-content metrics in Figure~\ref{fig:compaction-behavior} check both, yet a summary can satisfy them while proposing a next action incompatible with the state it records, violating what we call \emph{summary self-consistency}. Figure~\ref{fig:self-consistency-case} illustrates this gap through paraphrased summaries from AutoCompact-SFT and AutoCompact on the same SWE-bench Verified task.

The two summaries in Figure~\ref{fig:self-consistency-case} differ in exactly this respect. The AutoCompact-SFT summary records an unresolved syntax error but still proposes finishing without further action: the error is retained in the summary, yet it fails to constrain the proposed continuation. In contrast, the AutoCompact summary reports that the parameter and conditional-logic changes are complete and proposes verification before submission. This next step is consistent with the recorded implementation state. The corresponding AutoCompact-SFT run fails at evaluation with a \texttt{SyntaxError}, while the AutoCompact run passes the tests and resolves the task.

This contrast highlights summary self-consistency as a dimension of summary quality beyond information coverage. The difference may stem from the training objectives: SFT imitates demonstrated summaries, which does not ensure that the recorded state constrains the next action, whereas RL rewards a summary only through the success of the actions that follow it.

\section{Related Work}

\textbf{Coding agents.} Research on repository-level coding improves task-solving capabilities through both harness design and model training. Work on harness design organizes repository exploration, code editing, and validation through agent-computer interfaces, tool execution environments, and structured repair workflows~\citep{sweagent,openhands,xia2024agentless}. Work on model training improves coding and problem-solving capabilities through executable task environments, expanded task and trajectory datasets, and supervised or reinforcement learning~\citep{swegym,swesmith,swerl}. AutoCompact combines the two as model-harness co-design: the harness exposes a \texttt{compact()} action, and the model learns to use it jointly with task execution rather than through a separate compression module.

\textbf{Memory and context management.} Long-horizon agents manage growing interaction histories through memory hierarchies~\citep{memgpt}, abstraction of raw observations~\citep{synapse}, summaries of accumulated observations and interactions~\citep{acon,resum,contextasatool}, or compression of sub-trajectories and selected history spans~\citep{contextfolding,agentfold,swemem}. These works mainly design how retained context is represented. AutoCompact uses a simple representation, a single working-state summary alongside the original task and recent turns, and focuses on when to compact and how to continue afterward.

\textbf{Learning context-management policies.} Rubric- and guideline-based methods~\citep{selfcompact,acon} steer compaction through instructions rather than training the agent's own compaction behavior. Supervised methods learn from constructed trajectories~\citep{contextasatool,agentfold,swemem}, for example by inserting context-management actions into completed trajectories~\citep{contextasatool}, which keeps the original actions after each insertion, or by synthesizing them, as in concurrent work SWE-MeM~\citep{swemem}, where a stronger model decides when to compress a span of steps and writes the summary but generates no task actions. RL methods~\citep{resum,contextfolding,compactionrl,swemem} optimize context management through rewards, in SWE-MeM with additional step-level masks for memory-management failures, but do not directly supervise the actions that follow compaction. Unlike these methods, AutoCompact corrects the agent's own compaction decisions, summaries, and subsequent actions during data collection, executes the corrected outputs, and refines this behavior with outcome-based RL alone.

\section{Conclusion}

AutoCompact trains coding agents to manage context as part of task execution. Judge-corrected trajectories provide supervision for compaction decisions, summary construction, and the actions that follow compaction, followed by outcome-based reinforcement learning that jointly optimizes coding and compaction. Experiments on SWE-bench Verified and SWE-PolyBench Verified show improved task success across all evaluated inference budgets, and ignoring the compaction calls of the same trained model lowers pass rates, showing that the gains do not come from training alone. These findings highlight the value of training not only the construction of compacted states but also the agent's behavior that follows from these states, which prior methods leave unsupervised.

\textbf{Limitations and future work.} Due to resource constraints, RL training uses 32K-token sequences, shorter than the 256K context window used at evaluation. Nevertheless, the gains across all evaluated budgets in both the 256K and 16K settings indicate that this training setting suffices to learn useful proactive compaction. More broadly, AutoCompact is a form of model-harness co-design: the harness provides the compaction mechanism, while the model learns when to invoke it, what to preserve, and how to continue afterward. We study this co-design within a single scaffold. Widely used agent harnesses such as Codex and Claude Code currently compact the context automatically when it approaches the window limit, which is the length-triggered design discussed in Section~\ref{sec:intro}. Our 16K results suggest that combining such mechanisms with learned proactive compaction could further improve these systems, which we leave for future work.

\subsubsection*{AI Use Statement}
\label{sec:ai_use_statement}

In this work, we used generative AI tools for language editing and trajectory correction for training. We have reviewed all AI-assisted work. We take responsibility for the final content of this work, including text, claims, or artifacts produced with the aid of generative AI.

\bibliography{iclr2027_conference}
\bibliographystyle{iclr2027_conference}

\end{document}

%% file: math_commands.tex
\usepackage{amsmath,amsfonts,bm}

\def\eqref#1{equation~\ref{#1}}

\def\1{\bm{1}}

\DeclareMathAlphabet{\mathsfit}{\encodingdefault}{\sfdefault}{m}{sl}
\SetMathAlphabet{\mathsfit}{bold}{\encodingdefault}{\sfdefault}{bx}{n}

